\documentclass[10pt,twocolumn,letterpaper]{article}

\usepackage{wacv}
\usepackage{times}
\usepackage{epsfig}
\usepackage{graphicx}
\usepackage{amsmath}
\usepackage{array}
\usepackage{amssymb}
\usepackage{enumitem}
\usepackage{comment}
\usepackage{textcomp}
\newcommand*\rot{\rotatebox{90}}
\newcolumntype{P}[1]{>{\centering\arraybackslash}p{#1}}
\usepackage[normalem]{ulem}
\useunder{\uline}{\ul}{}
\usepackage{float}
\usepackage{cite}
\usepackage{balance}
\usepackage{lastpage}

\usepackage[pagebackref=true,breaklinks=true,letterpaper=true,colorlinks,bookmarks=false]{hyperref}

\wacvfinalcopy 

\def\wacvPaperID{85} 

\ifwacvfinal\fi
\begin{document}

\title{MLSL: Multi-Level Self-Supervised Learning for Domain Adaptation with Spatially Independent and Semantically Consistent Labeling}

\author{Javed Iqbal and Mohsen Ali \\
Information Technology University of Punjab, Lahore, Pakistan.\\
{\tt\small (javed.iqbal, mohsen.ali)@itu.edu.pk}
}


\maketitle
\ifwacvfinal\thispagestyle{empty}\fi

\begin{abstract}
Most of the recent Deep Semantic Segmentation algorithms suffer from large generalization errors, even when powerful hierarchical representation models based on convolutional neural networks have been employed.  
This could be attributed to limited training data and large distribution gap in train and test domain datasets.
In this paper, we propose a multi-level self-supervised learning model for domain adaptation of semantic segmentation.  
Exploiting the idea that an object (and most of the stuff given context) should be labeled consistently regardless of its location, we generate spatially independent and semantically consistent (SISC) pseudo-labels by segmenting multiple sub-images using base model and designing an aggregation strategy.
Image level pseudo weak-labels, PWL, are computed to guide domain adaptation by capturing global context similarity in source and domain at latent space level. Thus helping latent space learn the representation even when there are very few pixels belonging to the domain category (small object for example) compared to rest of the image.   
Our multi-level Self-supervised learning (MLSL) outperforms existing state-of-art (self or adversarial learning) algorithms. Specifically, keeping all setting similar and employing MLSL we obtain an mIoU gain of $5.1\%$ on GTA-V to Cityscapes adaptation and $4.3\%$ on SYNTHIA to Cityscapes adaptation compared to existing state-of-art method.

\end{abstract}
\vspace{-0.4cm}
\section{Introduction}
\vspace{-0.2cm}
\label{sec:intro}

With the evolution of deep learning methods during the last decade and the availability of densely labeled datasets \cite{Cordts2016Cityscapes, Ros_2016_CVPR, Richter_2016_ECCV}, a considerable attention has been devoted to improving the performance of semantic segmentation \cite{long2015fully, badrinarayanan2015segnet, chen2014semantic, noh2015learning, chen2018deeplab, zhao2017pyramid, csurka2008simple}. 
Significant reliance of real-time applications like autonomous vehicles \cite{kitty2012we}, bio-medical imaging \cite{ronneberger2015unet}, etc. over robust and accurate semantic segmentation step has also helped it gain prominence in current research. 
However, with the limited datasets for such a complex task (pixel-wise annotation), the state-of-the-art models have been reported to produce large generalization errors \cite{zou2018unsupervised, vu2019advent}.
This occurs naturally, because the train data may vary from test data (domain shift) in many aspects like illumination, visual appearance, camera quality, etc. 
It is time consuming and labor-intensive to densely label high resolution images covering all the domain variations. Modern computer graphics makes it easier to train deep models using synthetic images with computer generated dense labels \cite{Richter_2016_ECCV, Ros_2016_CVPR}. However, these simulated-scene datasets are significantly different in visual appearance and object structures compared to real-life road-scene datasets, limiting the model performance. 
To overcome these domain shift issues, many techniques have been proposed to adapt the target data distribution \cite{hoffman2016fcns, hoffman2017cycada, tsai2018learning}. Here our focus is to adapt the target domain dataset without labels in an unsupervised manner using Self-supervised learning.  

Due to large real-world applications, unsupervised domain adaptation (UDA) is a well-studied field in the current decade and aims to generalize to unseen data using only the labeled data of source domain.  
In UDA, most of the algorithms try to match the source and target data distribution using adversarial loss \cite{goodfellow2014generative} either at structured output level \cite{tsai2018learning} or latent space features level \cite{chen2017no, mancini2018boosting, chen2017road} respectively. Similarly, UDA based on adversarial learning augmented with other methods have recently produced good results on adaptation of semantic segmentation \cite{vu2019advent, zhang2018fully}. 
However, Zou et al. in \cite{zou2018unsupervised} showed that a comparative performance can be achieved using an alternative method contrary to adversarial learning with less computational resources required compared to these complex methods. They introduced a class balanced self-supervised training method by generating pseudo-labels using the source-data trained model and tried to minimize a single loss function. However, they failed to capture the global context of the image referenced to categories and also the generated pseudo-labels had high uncertainty.

 
 

In this work, we propose a novel Multi-level Self-Supervised learning (MLSL)  approach for UDA of semantic segmentation. 
The proposed approach consists of two complementary strategies.
First, we propose \textit{spatially independent and semantically consistent} (SISC) pseudo-labels generation process. We make reasonable assumption that an object should be segmented with similar label regardless of the location of the object. Same could be said about the stuff representing grass, road, sky, etc., given a reasonable context in surrounding.  
Using base model, multiple sub-images (extracted from an image) are segmented independently and output probability volume is aggregated.
This not only generates better pseudo-labels than single instance (SI) based ones, the assumption is more general than the spatial consistency assumption used by \cite{zou2018unsupervised}.  

Secondly, we enforce the global context and small object information preservation while adaptation by attaching a category based image classification module at latent space level. 
For each target image, Image level labels, called \textit{pseudo weak-labels} (PWL) are generated using SISC pseudo-labels and size statistics collected from source domain.
In summary, our main contributions are :
\begin{enumerate}[noitemsep,nolistsep]
\item A Multi-level self learning strategy for UDA of semantic segmentation by generating pseudo-labels at fine-grain pixel-level and image level, helping identify domain invariant features at both latent and output level. 
\item Designing a strategy, based on a reasonable assumption that for most categories labels should be location invariant given enough context, to generate \textit{spatially independent and semantically consistent} pixel-wise pseudo-labels 
\item Using category wise size statistics to help build PWL and train latent space. 
\item State-of-the-art performance on benchmark datasets by further augmenting the \textit{class spatial and category distribution priors}.
\end{enumerate}

\begin{figure*}[t]
 	\centering
 	\includegraphics[width=6.6in]{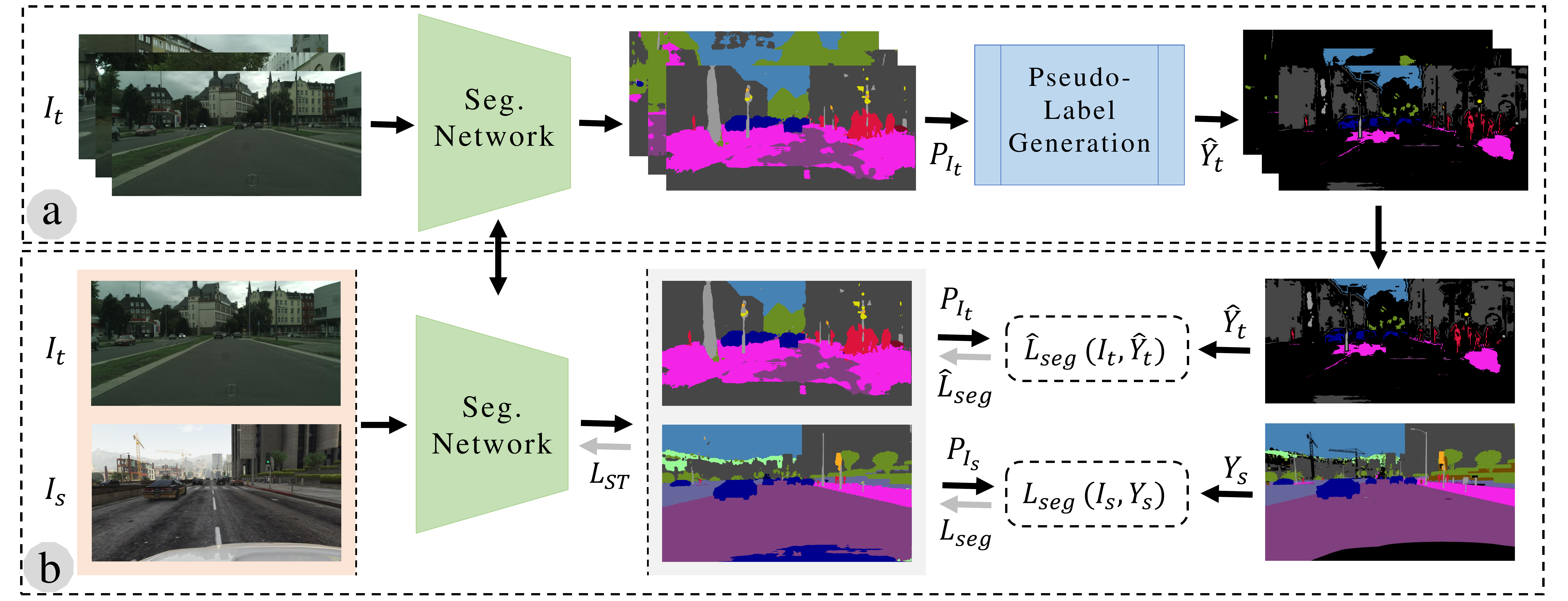}
 	\caption{An illustration of the alternating self-supervised learning method for UDA of semantic segmentation. (a) shows pseudo-label generation and (b) shows segmentation network training on source and target images. (a) and (b) are repeated iteratively.}
 	\label{img:1}
\vspace{-0.5cm}
\end{figure*}



\section{Related Work}
\label{sec:relatedWork}

Due to the evolution of deep learning methods, most of the computer vision tasks including but not limited to object detection, semantic segmentation, etc., are shifted to deep neural networks based methods \cite{chen2018domain}.
In \cite{long2015fully}, the authors proposed a fully convolutional network for pixel-level dense classification for the first time. Following them, many researchers proposed state-of-the-art methods for semantic segmentation taking the performance to an acceptable level for many computer vision tasks \cite{badrinarayanan2015segnet, chen2018deeplab, zhao2017pyramid}. 

Domain adaptation is a widely studied area in computer vision for segmentation, detection, and classification tasks. With the emergence of semantic segmentation algorithms \cite{badrinarayanan2015segnet, long2015fully, chen2014semantic}, availability of datasets \cite{Cordts2016Cityscapes, Ros_2016_CVPR, Richter_2016_ECCV} and modern applications demanding real-time constraints, e.g., self-driving cars, domain adaptation for semantic segmentation is in the spotlight. Many approaches exploited an appealing direction in semantic segmentation using domain adaptation from synthetic dataset to real-life datasets \cite{tsai2018learning, chen2017road}. 
The underlying idea of UDA include matching target and source features using discrepancy minimization \cite{zhang2018fully, mancini2018boosting}, self-supervised learning with pseudo-labels \cite{zou2018unsupervised, tri2018fully} and re-weighting source domain to look like target domain \cite{sankaranarayanan2018learning, hoffman2017cycada}. This work thoroughly investigates the unsupervised domain adaptation for semantic segmentation with focus on self-supervised learning approach. 


Adversarial learning is the most explored method for UDA of semantic segmentation \cite{chen2017road, chen2017no, tsai2018learning}. Adversarial loss-based training is exploited for feature matching, structured output matching, and re-weighting processes frequently in UDA. 
The authors in \cite{mancini2018boosting} and \cite{sankaranarayanan2018learning} exploited latent space representations and used an adversarial loss to match the latent space features of source and target domains. Similarly, Chen et al. \cite{chen2017no} used the adversarial loss for UDA of semantic segmentation augmented with class-specific adversaries to enhance the adaptation performance. 
The authors in \cite{zhang2018fully} also proposed the latent space domain matching based on adversarial loss augmented with appearance adaptation network at the input. They tried to combine the latent space representation adaptation and re-weighting process and observed a significant gain in performance. 
In \cite{hoffman2017cycada} the authors adapted similar approach to first transform the fully labeled source images to target images, train the segmentation model using the labeled source data, and then adapt further to target data. Rui et al. \cite{dlow_2019_CVPR} devised a domain flow approach to transfer source images to new domains using adversarial learning at intermediate levels. In \cite{structure_2019_CVPR}, the authors leveraged the spatial structure of source and target domain dataset, and working in latent space, proposed domain independent structure and domain specific texture based composite architecture for UDA.
However, due to high dimensional feature representation at latent space, it is hard to adapt to new data distributions using adversarial loss because of the instability of the adversarial learning process. 

In \cite{tsai2018learning}, the authors proposed a structured output domain adaptation based on adversarial learning. Their proposed method does not suffer from high dimensional representation of latent space and performs well due to a defined structure of road scene imagery at the output. They proposed state-of-the-art results in comparison with previous methods and also provided a baseline solution for other methods.
Zou et al. \cite{zou2018unsupervised} proposed a comparative performance method based on iterative learning. They proposed a class balanced self-training mechanism and obtained state-of-the-art performance using spatial priors in the pseudo-labels generation process. A tri-branch UDA model for semantic segmentation is proposed in \cite{tri2018fully}, where they generate pseudo-labels from two branches and train the third branch on that pseudo-labels alternatively.  
The authors in \cite{vu2019advent} stated that, only adversarial learning at latent space or output space is not enough to learn the target distribution. They used a direct entropy minimization algorithm augmented with an entropy-based adversarial loss for UDA of semantic segmentation.

In summary, the existing solutions are suffering due to various problems e.g. latent space adaptation suffers from high dimensional feature representation, output space adaptation struggles with small and thin objects, re-weighting independently is not enough to achieve the goal. Similarly, the existing iterative methods are not capable to generate good pseudo-labels and cannot capture the global image context. In this work we propose category-based image classification using PWL and SISC based self-supervised learning for domain adaptation of semantic segmentation. 
\begin{figure*}[t]
	\centering
	\includegraphics[width=\linewidth]{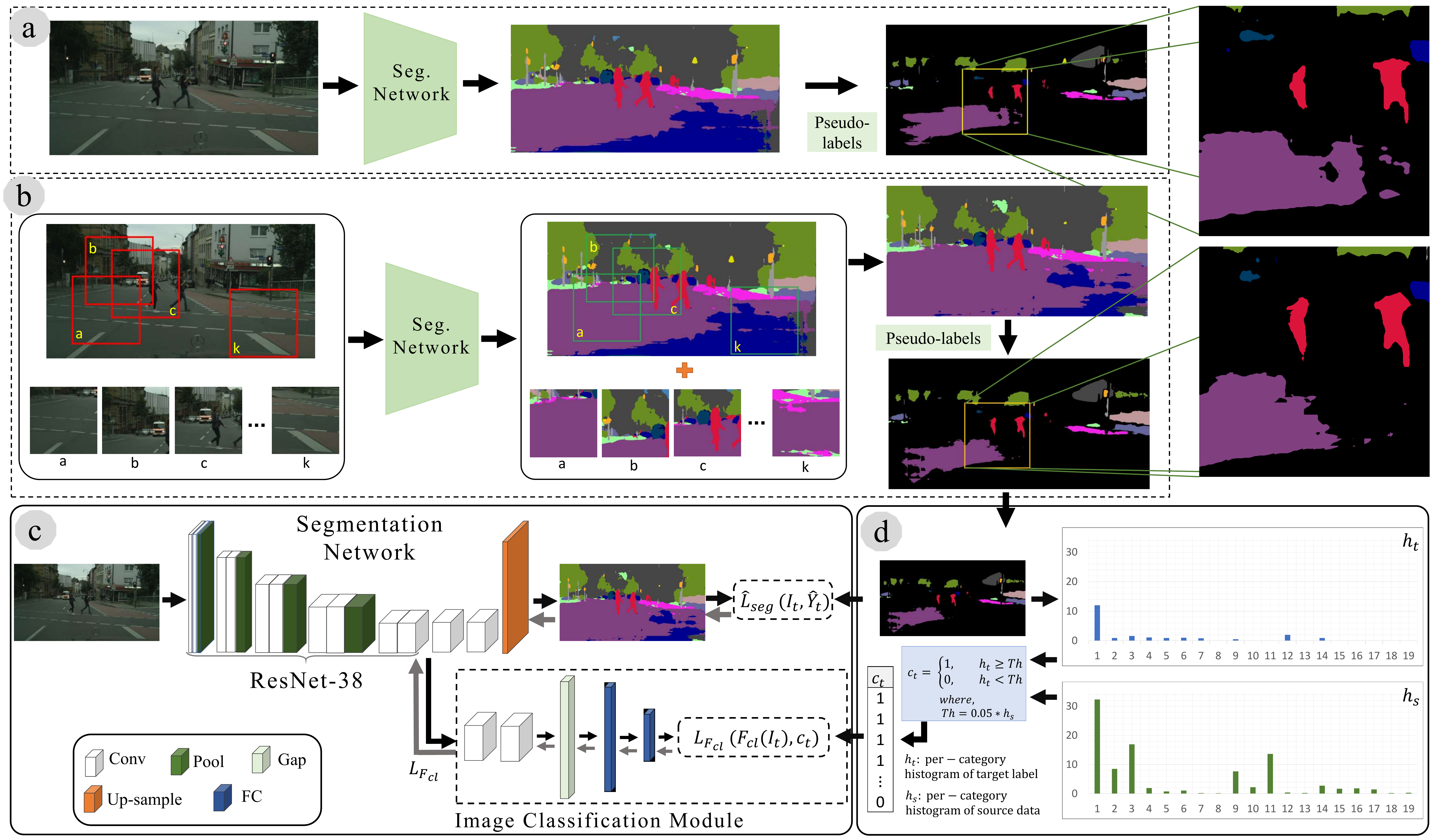}
	\caption{(a) Single-inference pseudo-label generation, (b) SISC pseudo-labels generation where, from left to right: patches are extracted randomly, segmented, recombined, normalized and pseudo-labels are generated. (c) shows the semantic segmentation and category-based image classification model, and (d) describes the PWL generation process.}
	\label{img:2}
	\vspace{-0.5cm}
\end{figure*}
\section{Approach}
\label{sec:method}
In this section, we present the proposed self-supervised and weakly-supervised learning approaches based on SISC pseudo-labels and PWL for domain adaptation of semantic segmentation. We start with existing state-of-the-art networks in semantic segmentation \cite{wu2019Resnet38} and self-training for domain adaptation \cite{zou2018unsupervised} as baseline methods and plugin additional modules for proposed approaches.
Fig. \ref{img:1} illustrates, iterative self-supervised learning technique for UDA.

\subsection{Preliminaries}
Let $I_s \in \mathbb{R} ^{H\times W\times 3}$ and $Y_s \in \mathbb{R} ^{H\times W\times C}$ where, $I_s$ corresponds to RGB images of source dataset with resolution $H\times W$ and $Y_s$ are ground truth labels as C-classes one-hot vectors with same spatial resolution as $I_s$. Let $G$ be a fully convolutional network which predicts softmax outputs $G(I) = G(I ^{H\times W\times 3}) = P_I^{H\times W\times C} = P_I$ for an input image $I$. One needs to learn the parameters $w_g$ of $G$ by minimizing the cross-entropy loss given in Eq. \ref{eqn:1} on source domain images.
\vspace{-0.2cm}
\begin{equation}
\small
    L_{seg} (I_s, Y_s) = -\sum_{H, W, C} Y_s^{H\times W\times C} log(P_{I_s}^{H\times W\times C})
\label{eqn:1}
\end{equation}
\vspace{-0.4cm}

If ground truth labels for target dataset are available, the most direct strategy would be to use Eq. \ref{eqn:1} and fine-tune the source trained model to target dataset. 
However, labels for target dataset are not available most of the time especially in real-time applications, e.g., self-driving cars. Therefore, an alternate way for unsupervised domain adaptation is to fine-tune the source trained model on the most confident outputs called ``pseudo-labels", which the model produces on target domain images. The pseudo-labels have exactly the same dimensions as $Y_s$. The loss function for the target domain images is formulated as follows:
\begin{equation}
\small
    \hat{L}_{seg} (I_t, \hat{Y}_t) = -\sum_{H, W, C} d^{H\times W} \hat{Y}_t^{H\times W\times C} log(P_{I_t}^{H\times W\times C})
\label{eqn:2}
\end{equation}
where $\hat{L}_{seg} (I_t, \hat{Y}_t)$ in Eq. \ref{eqn:2} is self-training loss with $\hat{Y}_t$ as the pseudo-labels one-hot vectors with $C$ classes, and $d^{H\times W}$ is a binary map, obtained from pseudo-labels $\hat{Y}_t$ e.g., $d_{ij} = 1$ if any pseudo-label is there at $\hat{Y}_{t_{ij}}$, and $d_{ij} = 0$ if there is no pseudo-label assigned at $\hat{Y}_{t_{ij}}$, where $i=1,...,H$ and $j=1,...,W$.  $d$ allows to back propagate loss for those pixel locations only, which are assigned pseudo-labels. We name the training method as ``self-supervised learning" or ``self-training".

\subsection{Semantically consistent pseudo-labels}
\label{sec:sisc}
Training a network using single inference (SI) generated pseudo-labels only misleads the training process as there is no guarantee over the quality of pseudo-labels. An initial optimal strategy is to jointly train the segmentation network using the ground truth labels of source images and the generated pseudo-labels of target images. The joint loss function is given by Eq. \ref{eqn:3}.
\begin{equation}
\small
   \underset{w_g} {min} L_{ST} ( I_s, Y_s, I_t, \hat{Y}_t) = L_{seg} (I_s, Y_s) + \hat{L}_{seg} (I_t, \hat{Y}_t)
\label{eqn:3}
\end{equation}
where, $L_{seg} (I_s, Y_s)$ is the loss of source images  and $\hat{L}_{seg} (I_t, \hat{Y}_t)$ is the loss of target images  given in Eq. \ref{eqn:1} and Eq. \ref{eqn:2} respectively. To minimize the loss in Eq. \ref{eqn:3}, we follow the two stage alternative process given below: 
\begin{enumerate}[noitemsep ,nolistsep]
\item Generate pseudo-labels by fixing the model parameters $w_g$.

\item Minimize the loss in Eq. \ref{eqn:3} with respect to $w_g$ by fixing the pseudo-labels $\hat{Y}_t$ generated in the previous step. 
\end{enumerate}

In this work, Step 1 and Step 2 are executed alternatively and repeated for multiple iterations. A work-flow of the proposed algorithm is shown in Fig. \ref{img:1}. Step 1 tries to generate pseudo-labels using the output softmax probabilities of the target images based on the more confident examples. Once the pseudo-labels are generated, Step 2 updates the model parameters $w_g$ using stochastic gradient descent (SGD) by minimizing the loss function given in Eq. \ref{eqn:3}. 

\textbf{Spatially independent and semantically consistent pseudo-labels:} Instead of generating pseudo-labels using SI, (e.g., segmenting the whole image simultaneously), we generate ``spatially independent and semantically consistent (SISC)" pseudo-labels. We leverage the spatial independence of our baseline semantic segmentation model to generate spatially independent and semantically consistent predictions. 
To quantitatively show the contribution of semantic consistency, we evaluate the softmax predictions based on different spatial context and select the most consistent ones. For each target image $I_t$, we select $K$ partially overlapping patches $[{p_1, p_2,...,p_K}]$ of size $h \times w$ each. Each patch $p_i$ is passed through the segmentation algorithm to assign pixel-wise confidence vectors using softmax outputs. 
The output softmax probabilities for each patch are added to an empty matrix $P_{I_c} \in \mathbb{R} ^{H\times W\times C}$ in specific locations where each patch belongs, and generate the composite output. Each pixel in $P_{I_c}$ has an associated count based on the number of occurrences in different patches during inference. We normalize $P_{I_c}$ with associated counts to obtain a normalized probability map and forward it to pseudo-label selection step which chooses the most confident outputs as pseudo-labels. The whole process of patch-based and single inference based pseudo-label generation is shown in Fig. \ref{img:2}. 


Unlike simple pseudo-labels generation methods which suffer from category distribution imbalance problem, we use the category-balanced pseudo-label selection similar to the method used in \cite{zou2018unsupervised}. Using the obtained normalized probability map, we further normalize the category-wise probabilities and select the pixels having high probability within a specific category. For example, we select all pixels locations which are assigned to be ``road", normalize probabilities on that locations and then select the most confident ones. 
This process balances the inter-category pseudo-labels ratio and avoids the training process to adapt simple examples only. The obtained pseudo-labels belong to the more consistent pixels inferred without the global view. The loss function given in Eq. \ref{eqn:3} is minimized using the original labels for source domain and SISC pseudo-labels for the target domain. 

\subsection{Pseudo weak-labels guided domain adaptation}
\label{sec:pwl}

The cross-entropy loss for an input image/label pair defined in Eq. \ref{eqn:1} calculates the sum of independent pixel-wise entropies, dealing with each pixel and label at the location independently. 
Thus ignoring any spatially global information, prone to effected by sparse erroneous pseudo-labels. 
Due to unbalanced pixels per category distribution, minimizing the summation of independent pixels entropies ignores the global data distribution. Even balancing the labels \cite{zou2018unsupervised}, the low-density classes fades (for target domain) as self-training proceeds. 

We employ the pseudo weak-labels (PWL), guided multi-task weakly-supervised learning to regularize the pixel wise cross-entropy loss. 
The PWL based category level cross-entropy loss is attached at the encoder level while adapting. This forces the latent space to learn to represent target categories, even for the small objects whose latent space representation might be faded if only pixel-wise cross-entropy loss is used. 

\subsubsection{PWL Filtering}
\label{sec:pwlFiltering}
The pixel-wise pseudo-labels are too noisy to generate the image level pseudo-labels. 
Assuming that source and target have similar objects and their instances, we build a naive model for the category's size relationship with the image. 
From the source dataset we calculate $h_s = \{ m_1, m_2, \dots, m_c\}$, to represent mean size of each class, where
\begin{equation}
\footnotesize
m_i = \frac{1}{(\sum_{j=1}^{N}1_{i}^{j}) \times H \times W} \sum_{j=1}^{N} 1_{i}^{j}  \{\sum_{x=1}^{H}\sum_{y=1}^{W} Y_s^j(x,y,i)\}
\label{eqn:4}
\end{equation}
$N$ stands for total images, and indicator function $\mathbf{1}_i^j$ is 1 if $j^{th}$ image has class $i$, otherwise zero.
For each target image $I_t$, we compute SISC pseudo-labels $\hat{Y}_t$ and use it to compute array $h_t$. PWL vector for image $I_t$ is an indicator vector $c^{pwl}$, s.t. $c^{pwl}_i=1$ if $h_t(i)>\eta h_s(i)$ otherwise zero.   $\eta$ is a small value chosen by the user.

\subsubsection{PWL Loss}
Given any image $I$, an image classification module $F_{cl}$ is designed to input the latent space representation (in this case of ResNet-38), and predict labels (Fig. \ref{img:2}(c)). Instead of softmax, we use sigmoid so that it can predict multiple labels for the image and use binary cross-entropy loss function given in Eq.\ref{eq:binCE} .
\begin{equation}
\label{eq:binCE}
\footnotesize
L_{F_{cl}} (I, c) = - \frac{1}{C} \sum_{i=1}^C (c_i) log(F_{cl}(I)) + (1 - c_i) log(1 - F_{cl}(I))
\end{equation}
For the source images $I_s$, indicator vector $c$ represents image level label crated from ground truth segmentation labels. For the images in target domain $I_t$, image level weak-labels $c^{pwl}$ are created as detailed in Sec. \ref{sec:pwlFiltering}. 


\subsection{Final Loss Function}

The overall loss function for segmentation network and category-based image classification network for source domain is the composition of both \ref{eqn:1} and \ref{eq:binCE}, and is given by 
\begin{equation}
\small
L_{cmp} (I, Y, c) = L_{seg} (I, Y) + \lambda_{F_{cl}} L_{F_{cl}} (I, c)
\label{eqn:5}
\end{equation}
where $\lambda_{F_{cl}}$ is the scaling factor and $c$ is image level label. 
The combined loss function for self-supervised and weakly-supervised learning is given by;
\begin{equation}
\small
   L_{STWL} ( I_s, Y_s, I_t, \hat{Y}_t, c) = L_{cmp} (I_s, Y_s, c)\\ + \hat{L}_{cmp} (I_t, \hat{Y}_t, c)
\label{eqn:7}
\end{equation}
Eq. \ref{eqn:7}, is minimized using criteria described in Sec. \ref{sec:sisc}.
\section{Experiments}
\label{sec:exp}
In this section, we present experimental details and discuss the main results of our proposed UDA methods.

\begin{table*}[h]
\centering
\caption{Semantic segmentation performance when the model trained on GTA-V dataset is adapted to Cityscapes dataset. We present the results of our proposed SISC pseudo-labels based self-supervised learning and PWL augmented self-training. We use the competitive baseline model and show a thorough comparison with existing state-of-the-art methods. The abbreviations "ST" and "Adv" indicates the self-training (self-supervised learning) and adversarial learning respectively.}
\resizebox{\textwidth}{!}{
\begin{tabular}{l|c|ccccccccccccccccccc|c}
\hline 
\multicolumn{22}{c}{GTA-V $\rightarrow$ Cityscapes}\\
\hline
Methods     & \rot{Appr.} & \rot{Road}  & \rot{Sidewalk} & \rot{Building} & \rot{Wall}  & \rot{Fence} & \rot{Pole}  & \rot{T. Light} & \rot{T. Sign} & \rot{Veg.} & \rot{Terrain} & \rot{Sky}   & \rot{Person} & \rot{Rider} & \rot{Car}   & \rot{Truck} & \rot{Bus}   & \rot{Train} & \rot{M.cycle} & \rot{Bicycle} & \rot{mIoU}  \\ \hline \hline
ResNet-38 \cite{wu2019Resnet38}      & -        & 70.0          & 23.7          & 67.8          & 15.4          & 18.1          & {\ul 40.2}    & 41.9          & 25.3          & 78.8          & 11.7          & 31.4          & {\ul 62.9}    & {\ul 29.8}          & 60.1          & 21.5          & 26.8        & 7.7           & 28.1          & 12.0          & 35.4          \\
AdaptSetNet \cite{tsai2018learning}    & Adv      & 86.5          & 36.0          & 79.9          & 23.4          & 23.3          & 23.9          & 35.2          & 14.8          & 83.4          & {\ul 33.3}    & 75.6          & 58.5          & 27.6          & 73.7          & 32.5          & 35.4        & 3.9           & 30.1          & 28.1          & 42.4          \\
Saleh et al  \cite{saleh2018effective}   & ST       & 79.8          & 29.3          & 77.8          & 24.2          & 21.6          & 6.9           & 23.5          & {\ul 44.2}    & 80.5          & \textbf{38.0} & 76.2          & 52.7          & 22.2          & 83.0          & 32.3          & 41.3        & 27.0          & 19.3          & 27.7          & 42.5          \\
MinEnt \cite{vu2019advent}         & ST      & 86.2          & 18.6          & 80.3    & 27.2 & 24.0          & 23.4          & 33.5          & 24.7          & \textbf{83.3} & 31.0          & 75.6          & 54.6          & 25.6    & 85.2          & 30.0          & 10.9        & 0.1           & 21.3          & 37.1          & 42.3          \\
DLOW  \cite{dlow_2019_CVPR}          & Adv      & 87.1          & 33.5          & {\ul 80.5}          & 24.5          & 13.2          & 29.8          & 29.5          & 26.6          & 82.6          & 26.7          & {\ul 81.8}    & 55.9          & 25.3          & 78.0          & {\ul 33.5}    & 38.7        & 0.0           & 22.9          & 34.5          & 42.3          \\
CLAN \cite{clan_2019_CVPR}           & Adv      & 87.0          & 27.1          & 79.6          & {\ul 27.3}          & 23.3          & 28.3          & 35.5          & 24.2          & 83.6          & 27.4          & 74.2          & 58.6          & 28.0          & 76.2          & 33.1          & 36.7        & 6.7           & 31.9          & 31.4          & 43.2          \\
All Structure \cite{structure_2019_CVPR}  & Adv      & \textbf{91.5} & 47.5          & \textbf{82.5} & 31.3          & 25.6          & 33.0          & 33.7          & 25.8          & 82.7          & 28.8          & \textbf{82.7} & 62.4          & \textbf{30.8} & 85.2          & 27.7          & 34.5        & 6.4           & 25.2          & 24.4          & 45.4          \\
CBST-SP \cite{zou2018unsupervised}        & ST       & 88            & \textbf{56.2} & 77            & \textbf{27.4}    & 22.4          & \textbf{40.7} & \textbf{47.3} & 40.9          & 82.4          & 21.6          & 60.3          & 50.2          & 20.4          & 83.8          & 35            & \textbf{51} & 15.2          & 20.6          & 37            & 46.2          \\ \hline
Ours (SISC)     & ST       & {\ul 91.0}          & {\ul 49.3}    & 79.9          & 24.4          & \textbf{27.9} & 37.9          & 45.1          & \textbf{45.1} & 81.3          & 19.0          & 61.7          & \textbf{63.9} & 28.0          & \textbf{86.5} & 23.9          & 42.3        & \textbf{41.9} & {\ul 33.1}    & {\ul 44.4}    & {\ul 48.7}    \\
Ours (SISC-PWL) & ST       & 89.0    & 45.2          & 78.2          & 22.9          & {\ul 27.3}    & 37.4          & {\ul 46.1}    & 43.8          & {\ul 82.9}    & 18.6          & 61.2          & 60.4          & 26.7          & {\ul 85.4}    & \textbf{35.9} & {\ul 44.9}  & {\ul 36.4}    & \textbf{37.2} & \textbf{49.3} & \textbf{49.0} \\ \hline

\end{tabular}
}
\label{table:1}
\vspace{-0.5cm}
\end{table*}

\subsection{Experimental setup}
\subsubsection{Datasets}
We follow the \textit{synthetic-to-real} setup for unsupervised domain adaptation. We use GTA-V \cite{Richter_2016_ECCV} and SYNTHIA \cite{Ros_2016_CVPR} as our source domain synthetic datasets and Cityscapes \cite{Cordts2016Cityscapes} as real-world target domain dataset. 
GTA-V consist of 24966 synthetic frames of spatial resolution $1052 \times 1914$ extracted from a video game. All the 24966 frames have pixel level labels available for 33 categories, but we used 19 categories compatible with real-world Cityscapes dataset. 
Similarly, we use SYNTHIA-RAND-CITYSCAPES set having 9400 synthetic frames of size $760 \times 1280$ from SYNTHIA dataset. We train and evaluate our baseline and proposed models with 16 common classes in SYNTHIA and Cityscapes. We also report the 13 classes evaluation as described in \cite{vu2019advent} and \cite{zou2018unsupervised}.

In both the experiments, we use the Cityscapes training set without labels for unsupervised domain adaptation and evaluate the adapted models on Cityscapes separate validation set having 500 images. We use standard mean Intersection-over-Union (mIoU) as our evaluation metric.
\vspace{-0.2cm}
\subsubsection{Model architecture}
\vspace{-0.1cm}

We use ResNet-38 \cite{wu2019Resnet38} as our baseline semantic segmentation model. The pre-trained ResNet-38 (trained on ImageNet \cite{russakovsky2015imagenet}) is trained for semantic segmentation on GTA-V and SYNTHIA datasets. 
The architecture of ResNet-38 contains 7-blocks are there followed by two segmentation layers and an upsampling layer. We also call the ResNet-38 as encoder for segmentation network and refer its output as latent space representation. The two convolution layers comprises of $3 \times 3$ filters with depth of 512 and $C$ (number of classes to segment). At the end the upsampling layer up-scales the output using bi-linear interpolation. 

Similarly, the image classification part discussed in Section \ref{sec:pwl} is a category (object/stuff) based image classification module augmented with ResNet-38. The image classification module consist of two convolution layers with filters $[1 \times 1, 3 \times 3]$ with depth $2048$ each. A global average pooling (GAP) layer is applied to capture the global nature of the feature map channels. The output of GAP is passed through two fully connected layers of depth $512$ and $C$ respectively. Relu activation function is applied except the last layer where sigmoid is used. 
\vspace{-0.4cm}
\subsubsection{Implementation and training details}
\vspace{-0.1cm}

We use MxNet \cite{mxnet15} deep learning framework and a single Core-i5 machine with 32GB RAM and a GTX 1080 GPU with 8GB of memory to implement the proposed methods for domain adaptation of semantic segmentation. Our proposed model uses SGD optimizer for training with an initial learning rate of $1 \times 10^{-4}$. To generate SISC pseudo-labels, $K=50$ is chosen (e.g. 50 sub-images of a target image are selected randomly).
For SISC pseudo-labels based self-supervised learning, a batch size of 2 is chosen while the weakly-supervised setup described in section \ref{sec:pwl} processes a single image only. To optimize the joint loss function given in Eq. ref{eqn:7}, the value of $\lambda_{F_{cl}}$ is investigated thoroughly (as shown in Section \ref{sec:abl}) and chosen as 0.025 to limit the image classification loss to back propagate large gradients. $\lambda_{F_{cl}}$ also controls the speed of adaptation with trade-off to segmentation performance, so the mentioned nominal value is used for all followed experiments. 

\begin{table*}[h]
\centering
\caption{Semantic segmentation performance of Cityscapes validation set when adapted from SYNTHIA dataset. 
We present mIoU and mIoU* (13-categories) comparison with existing state-of-the-art methods for Cityscapes validation set.
}
\resizebox{6.5in}{!}{
\begin{tabular}{l|c|cccccccccccccccc|c|c}
\hline 
\multicolumn{19}{c}{SYNTHIA $\rightarrow$ Cityscapes}\\
\hline
Methods     & \rot{Appr.} & \rot{Road}  & \rot{Sidewalk} & \rot{Building} & \rot{Wall}  & \rot{Fence} & \rot{Pole}  & \rot{T. Light} & \rot{T. Sign} & \rot{Veg.} &  \rot{Sky}   & \rot{Person} & \rot{Rider} & \rot{Car}    & \rot{Bus}    & \rot{M.cycle} & \rot{Bicycle} & \rot{mIoU} & \rot{mIoU*}  \\ \hline \hline
ResNet-38 \cite{wu2019Resnet38}  & -        & 32.6 & 21.5     & 46.5     & 4.81 & 0.03  & 26.5 & 14.8          & 13.1         & 70.8       & 60.3 & 56.6   & 3.5   & 74.1 & 20.4 & 8.9        & 13.1    & 29.2 & 33.6  \\
Road \cite{chen2017road}   & Adv      & 77.7 & 30.0     & 77.5     & 9.6  & 0.3   & 25.8 & 10.3          & 15.6         & 77.6       & 79.8 & 44.5   & 16.6  & 67.8 & 14.5 & 7.0        & 23.8    & 36.2 & 41.8  \\
AdaptSetNet \cite{tsai2018learning}  & Adv      & {\ul 81.7} & {\ul 39.1}     & {\ul 78.4}     & 11.1 & 0.3   & 25.8 & 6.8           & 9.0          & 79.1       & 80.8 & 54.8   & 21.0  & 66.8 & \textbf{34.7} & 13.8       & 29.9    & 39.6 & 45.8  \\
MinEnt \cite{vu2019advent}       & ST      & 73.5 & 29.2     & 77.1     & 7.7  & 0.2   & 27.0 & 7.1           & 11.4          & 76.7       & \textbf{82.1} & 57.2   & {\ul 21.3}  & 69.4 & 29.2 & 12.9       & 27.9    & 38.1 & 44.2  \\
CLAN  \cite{clan_2019_CVPR}        & Adv      & 81.3 & 37.0     & \textbf{80.1}     & -    & -     & -    & 16.1          & 13.7         & 78.2       & {\ul 81.5} & 53.4   & 21.2  & 73.0 & {\ul 32.9} & {\ul 22.6}       & 30.7    & -    & 47.8  \\
All Structure \cite{structure_2019_CVPR} & Adv      & \textbf{91.7} & \textbf{53.5}     & 77.1     & 2.5  & 0.2   & 27.1 & 6.2           & 7.6          & 78.4       & 81.2 & 55.8   & 19.2  & 82.3 & 30.3 & 17.1       & 34.3    & 41.5 & 48.7  \\
CBST  \cite{zou2018unsupervised}     & ST       & 53.6 & 23.7     & 75.0     & 12.5 & 0.3   & {\ul 36.4} & 23.5          & {\ul 26.3}         & 84.8       & 74.7 & {\ul 67.2}   & 17.5  & \textbf{84.5} & 28.4 & 15.2       & \textbf{55.8}    & 42.5 & 48.4  \\ \hline
Ours (SISC)   & ST       &73.7	&34.4	&78.7	&{\ul 13.7}	&\textbf{2.9}	&\textbf{36.6}	&{\ul 28.2}	&22.3	&{\ul 86.1}	&76.8	&65.3	&20.5	&81.7	&31.4	&13.9	&47.3	&{\ul 44.4}	&{\ul 50.8} \\
Ours (SISC+PWL)   & ST       &59.2	&30.2	&68.5	&\textbf{22.9}	&{\ul 1.0}	&36.2	&\textbf{32.7}	&\textbf{28.3}	&\textbf{86.2}	&75.4	&\textbf{68.6}	&\textbf{27.7}	&{\ul 82.7}	&26.3	&\textbf{24.3}	&{\ul 52.7}	&\textbf{45.2}	&\textbf{51.0} \\ \hline
\end{tabular}
}
\label{table:2}
\vspace{-0.5cm}
\end{table*}

\subsection{Experimental results}
The experimental results of our proposed approaches compared to baseline ResNet-38 and existing state-of-the-art UDA methods are presented in this section. Our proposed approaches perform superior to other methods for domain adaptation and produce state-of-the-art results on two benchmark datasets. We also describe in detail, the behaviour of proposed approaches when exploited with different settings and different source datasets.

\textbf{GTA-V to Cityscapes:} Table \ref{table:1} details the experimental results of 19 categories when adapted from GTA-V to Cityscapes. We use standard mIoU as semantic segmentation performance measure and report results on Cityscapes validation set. Our proposed approach of self-supervised learning with SISC pseudo-labels, shows state-of-the-art performance with ResNet-38 segmentation model. 
The SISC approach outperforms the latest approaches for UDA of semantic segmentation. 
Compared to MinEnt \cite{vu2019advent} which tries to minimize the self-entropy using direct entropy minimization, our SISC approach shows $13.1\%$ improvement in overall mIoU. Similarly, compared to the self-training approach presented in \cite{zou2018unsupervised}, the proposed SISC method outperforms it with a margin of $5.1\%$ in mIoU. 

Our weak-labels guided UDA approach tries to capture the global image context by category (object/stuff) based image classification. This model helps improving the overall performance, and especially boost the performance for small and less occurring objects as shown in Table \ref{table:1}. The consistency and accuracy of pseudo weak-labels for image classification enable this approach to help the segmentation model for better performance. With ResNet-38 baseline, pseudo weak-labels when combined with CBST \cite{zou2018unsupervised} provides $2.3\%$ boost in mIoU compared to simple CBST. Similarly, when SISC is augmented with PWL based image classification, the mIoU performance increases by $5.7\%$ from existing stat-of-the-art CBST-SP \cite{zou2018unsupervised} as shown in Table \ref{table:1}. The ensemble of the two proposed approaches for UDA achieve 49.0 mIoU on Cityscapes validation set, which sets a new benchmark. The high boost in performance shows that both the approaches are capable to extract domain independent representations and produce better segmentation results comparatively. 

\begin{figure*}[t]
	\centering
	\includegraphics[width=6.7in]{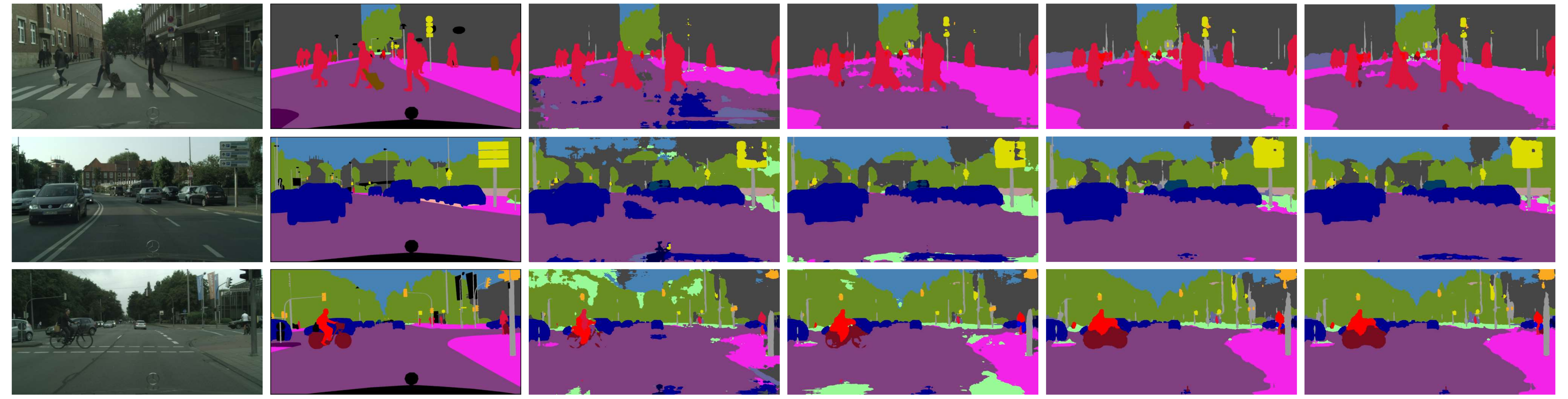}\\
	\footnotesize
	\begin{tabular}{P{2.6cm}P{2.4cm}P{2.5cm}P{2.5cm}P{2.45cm}P{2.6cm}}
    Target Image & Gound Truth & ResNet-38 \cite{wu2019Resnet38} &CBST-SP \cite{zou2018unsupervised} & Ours (SISC) & Ours (SISC+PWL)
    \end{tabular}
	\caption{Segmentation results on Cityscapes validation set when adapted from GTA to Cityscapes.}
	\label{img:4}
	\vspace{-0.2cm}
\end{figure*}

For a more fair comparison with other UDA methods, in Table \ref{table:3}, we show the mIoU gain with respect to specific baselines methods used. Compared to more complex models with very deep backbones, our approaches produces a higher gain of +13.6 points to source model surpassing the existing methods by a minimum margin of $20\%$.
Fig. \ref{img:4} shows some examples of semantic segmentation before and after domain adaptation. As illustrated in the figure, the segmentation results improves significantly with SISC and SISC+PWL based approaches compared to source and CBST-SP methods.  

\vspace{-0.2cm}
\begin{table}[H]
\footnotesize
\caption{Performance (mIoU, mIoU*) gain comparison between the GTA-V and SYNTHIA trained source models and the respective adapted models from GTA-V and SYNTHIA to Cityscapes.}
\centering
\begin{tabular}{p{2.3cm}|P{0.55cm}P{0.5cm}P{0.5cm}|P{0.55cm}P{0.5cm}P{0.5cm}}
\hline
Dataset         & \multicolumn{3}{c}{GTA $\rightarrow$ Cityscapes}          & \multicolumn{3}{|c}{SYN $\rightarrow$ Cityscapes}  \\
\hline
Methods         & Source only & UDA Algo.    & mIoU gain      & Source only & UDA Algo. & mIoU* gain  \\ \hline
\hline
FCN in the wild \cite{hoffman2016fcns}& 21.2     & 27.1          & 5.9             & 23.6     & 25.4       & 1.8        \\
Curriculam DA \cite{curr2017_ICCV}  & 22.3     & 28.9            & 6.6           & 28.4     & 34.82      & 6.42       \\
AdaptSetNet \cite{tsai2018learning}    & 36.6        & 42.4             & 5.8             & 38.6     & 46.7       & 8.1        \\
MinEnt \cite{vu2019advent}         & 36.6     & 42.3          & 5.7           & 38.6     & 44.2       & 5.6          \\
CLAN \cite{clan_2019_CVPR}            & 36.6        & 43.2             & 6.6             & 38.6     & 47.8       & 9.2        \\
All Structure \cite{structure_2019_CVPR}  & 36.6     & 45.4          & 8.8          & 38.6     & 48.7       & 10.1       \\
CBST \cite{zou2018unsupervised}            & 35.4     & 46.2          & 10.8          & 33.6     & 48.4       & 14.8       \\ \hline
Ours (SISC)     & 35.4     & {\ul 48.7} & {\ul 13.3} & 33.6     & {\ul 50.8}          & {\ul 17.2} \\ 
Ours (SISC+PWL)     & 35.4     & \textbf{49} & \textbf{13.6} & 33.6     & \textbf{51.0}          & \textbf{17.4} \\ \hline
\end{tabular}%
\label{table:3}
\vspace{-0.2cm}
\end{table}

\textbf{SYNTHIA to Cityscapes:} SYNTHIA is a more diverse dataset with multiple viewpoints and different spatial constraints compared to GTA-V and Cityscapes. In Table \ref{table:2}, we present the unsupervised adaptation results on Cityscapes validation set when adapted from SYNTHIA. The categories in SYNTHIA and Cityscapes do not fully overlap, so we have selected the common 16 classes as done in \cite{hoffman2016fcns, curr2017_ICCV, zou2018unsupervised} for evaluation. We have also reported the performance (mIoU*) over the 13 common classes as used in \cite{zou2018unsupervised, tsai2018learning, clan_2019_CVPR}. 
With ResNet-38 as baseline netwrok, our proposed SISC based sef-supervised learning method performs superior to existing state-of-the-art methods as shown in Table \ref{table:2}. Compared to MinEnt \cite{vu2019advent} which uses similar entropy minimization technique, our SISC based UDA approach achieves $14.2\%$ gain in mIoU and $13.3\%$ gain in mIoU*. Similarly, compared to CBST presented in \cite{zou2018unsupervised}, our SISC based approach gains $4.3\%$ and $4.7\%$ points in mIoU and mIoU* respectively.  
Our proposed PWL guided UDA approach combined with SISC based self-supervised learning provides $6.0\%$ and $5.1\%$ boost in mIoU and mIoU* respectively when compared with CBST. Compared to an ensemble method (adversarial training and self-training) \cite{vu2019advent}, our composite UDA method achieves $9.8\%$ and $7.1\%$ gain in mIoU and mIoU* respectively.

\begin{figure*}[t]
	\centering
	\includegraphics[width=6.7in]{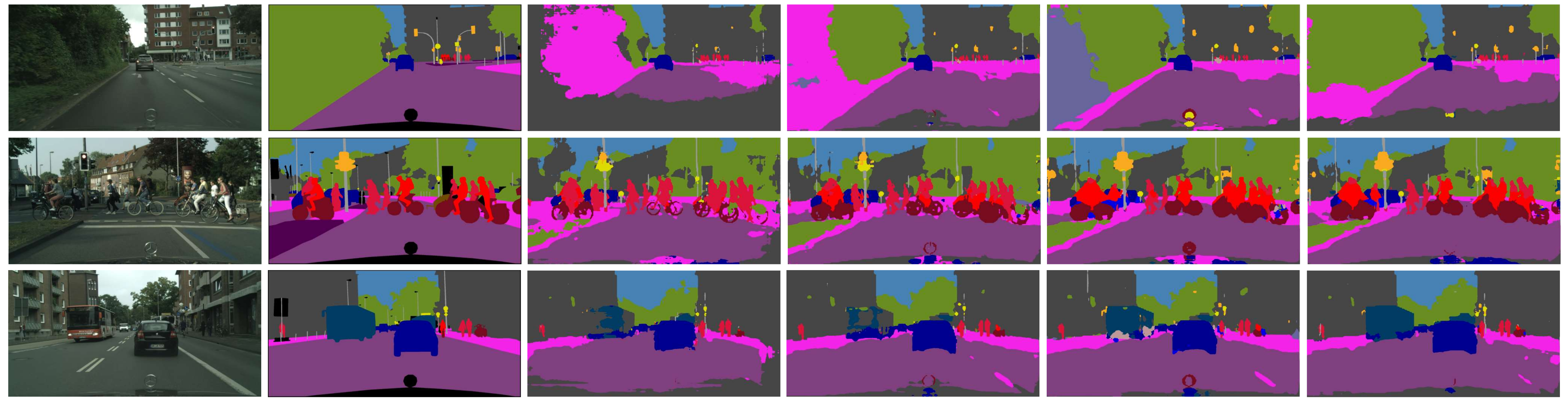} \\
	\footnotesize
	\begin{tabular}{P{2.6cm}P{2.4cm}P{2.5cm}P{2.5cm}P{2.45cm}P{2.6cm}}
    Target Image & Gound Truth & ResNet-38 \cite{wu2019Resnet38} &CBST-SP \cite{zou2018unsupervised} & Ours (SISC) & Ours (SISC+PWL)
    \end{tabular}
	\caption{Segmentation results on Cityscapes validation set when adapted from SYNTHIA to Cityscapes.}
	\label{img:5}
	\vspace{-0.6cm}
\end{figure*}

To make a more fair comparison with existing methods, Table \ref{table:3} shows the baseline, after adaptation, and gain in terms of mIoU*. It is fair to say, that our proposed methods outperforms the existing state-of-the-art methods achieving the gain over baseline with a minimum margin of $16.3\%$. 
In Fig. \ref{img:5}, some examples of semantic segmentation before and after UDA are shown. As illustrated, the segmentation results improves significantly with SISC and SISC+PWL based approaches compared to source and CBST methods.

\vspace{-0.1cm}
\subsection{Ablation experiments}
\label{sec:abl}
\vspace{-0.1cm}
\textit{Relative frequency based pseudo-labels:} 
Besides the adapted methodology in Section \ref{sec:sisc}, we also generated pixel classification relative frequency based pseudo-labels. The randomly selected patches like SISC are segmented and recombined in the large output map. A count is made for each pixel with respect to assigned category in each patch, and then relative frequency is calculated. This relative frequency is used as prediction probability and incorporated in pseudo-labels generation. Due to hard decision, the pseudo-labels generated were not effective and lead to a decline in the performance. 
\vspace{-0.2cm}
\begin{table}[H]
\scriptsize
\centering
\caption{Influence of $\lambda_{F_{cl}}$ and $\eta$ on overall performance.}
\begin{tabular}{ccccc}
\hline
\multicolumn{5}{c}{GTA-V $\rightarrow$ Cityscapes} \\
\hline
$\lambda_{F_{cl}}$        & 0.1        & 0.05       & 0.025       & 0.001                \\
SISC+PWL           & 46.0       & 48.1       & 49.0        & 48.24   \\
\hline
$\eta$& 0.0  & 0.1        & 0.05       & 0.025                \\
SISC+PWL           &45.5    & 46.0       & 49.0       & 47.33   \\
\hline
\end{tabular}
\label{table:5}
\vspace{-0.3cm}
\end{table}
\textit{Patch size selection:} Our base models for semantic segmentation in both cases are trained on $500 \times 500$ random patches selected from the whole image randomly. Following that nominal size, we have chosen $512 \times 512$ as our patch size for pseudo-label generation. We also tried with $256 \times 256$ patch size but on high resolution Cityscapes images, these small image patches were not contributing. For patch size greater than $512 \times 512$ there were GPU memory limitations. Similarly, we selected 25, 50 and 100 patches per image randomly for SISC pseudo-labels generation. 25 patches were not enough to capture the high resolution Cityscapes images and 100 patches were taking the process very slow with negligible gain over 50 patches. Therefore, for all experiments, we have chosen 50 random patches per image.
\textit{Category based image classification loss weight:} Since image classification is added as a supporting module to segmentation network, the loss contribution by this module should also be limited. We tried multiple weight factors, and selected $\lambda_{F_{cl}} = 0.025$ (Table \ref{table:5}).
\textit{Pseudo-weak-label generation:} For category based image classification loss, the PWL are generated from segmentation pseudo-labels. Since it is difficult to set a minimum number of pixels limit for a category to be labeled as present in an image. Therefore, we exploited the category distribution of source datasets and assigned pseudo weak-labels to present categories based on source data distribution. For GTA-V to Cityscapes, we select a category to be labeled as present in an image if, it has more pixels compared to the $5\%$ of mean category pixels of the same category in the source dataset. 
A detailed comparison along with respective mIoU is shown in Table \ref{table:5}.

\vspace{-0.2cm}
\section{Conclusions}
\vspace{-0.2cm}
In this paper, we have proposed,  Multi-level self learning strategy (MLSL) for UDA of semantic segmentation by generating pseudo-labels at fine-grain pixel-level and image level, helping identify domain invariant features at both latent and output space.
Using a reasonable assumption that labels of objects and stuff should be same regardless of their location, we generate Spatially independent but Semantically Consistent Labels.
Image level labels, called pseudo weak-label (PWL)  are generated by learning the pixel-wise object size distribution in the source domain images and using it as consistency check over SISC pseudo-labels.
Binary cross-entropy loss using PWL enforces latent space to preserve the information about the objects, helping domain adapt for small objects. 
This multi-level pseudo-label generation for self-supervised learning, allows the network to learn domain-invariant features at different hierarchical levels. 
The rigorous experimentation demonstrates that the proposed SISC based self-supervised method alone outperforms the existing state-of-the-art algorithms on benchmark datasets: mIoU* improves from $46.2$ to $48.7$ and $48.4$ to $50.8$ on GTA-V $\&$ SYNTHIA to Cityscapes. This includes both,  ones using self-supervision or adversarial learning. 
Augmented with a PWL based image classification module, our proposed method further improves the performance, especially in the small objects. 
Effectiveness of SISC and PWL is highlighted by the substantial improvement of mean IOU over the base model, which is significantly more than previous state-of-methods.


{\small
\bibliographystyle{unsrt}
\bibliography{egbib}
}

\end{document}